\documentclass[conference]{IEEEtran}
\IEEEoverridecommandlockouts
\usepackage{cite}
\usepackage{amsmath,amssymb,amsfonts}
\usepackage{algorithmic}
\usepackage{graphicx}
\usepackage{textcomp}
\usepackage{xcolor}
\usepackage{booktabs}
\usepackage{multirow}
\usepackage{array}
\def\BibTeX{{\rm B\kern-.05em{\sc i\kern-.025em b}\kern-.08em
    T\kern-.1667em\lower.7ex\hbox{E}\kern-.125emX}}
\begin{document}

\title{Computer Vision for Real-Time Monkeypox Diagnosis on Embedded Systems\\
\thanks{NSF Grant OIA-1849243}
}

\author{\IEEEauthorblockN{1\textsuperscript{st} Jacob M. Delgado-López}
\IEEEauthorblockA{\textit{Department of Computer Science and Engineering} \\
\textit{University of Puerto Rico Mayagüez}\\
Mayagüez, Puerto Rico \\
jacob.delgado@upr.edu}
\and
\IEEEauthorblockN{2\textsuperscript{nd} Ricardo A Morell-Rodriguez}
\IEEEauthorblockA{\textit{Department of Computer Science and Engineering} \\
\textit{University of Puerto Rico Mayagüez}\\
Mayagüez, Puerto Rico \\
ricardo.morell1@upr.edu}
\and
\IEEEauthorblockN{3\textsuperscript{rd} Sebastián O Espinosa-Del Rosario}
\IEEEauthorblockA{\textit{Department of Computer Science and Engineering} \\
\textit{University of Puerto Rico Mayagüez}\\
Mayagüez, Puerto Rico \\
sebastian.espinosa@upr.edu}
\and
\IEEEauthorblockN{4\textsuperscript{th} Wilfredo E. Lugo-Beauchamp}
\IEEEauthorblockA{\textit{Department of Computer Science and Engineering} \\
\textit{University of Puerto Rico Mayagüez}\\
Mayagüez, Puerto Rico \\
wilfredo.lugo1@upr.edu}
}

\maketitle

\begin{abstract}
The rapid diagnosis of infectious diseases, such as monkeypox, is crucial for effective containment and treatment, particularly in resource-constrained environments. This study presents an AI-driven diagnostic tool developed for deployment on the NVIDIA Jetson Orin Nano, leveraging the pre-trained MobileNetV2 architecture for binary classification. The model was trained on the open-source Monkeypox Skin Lesion Dataset, achieving a 93.07\% F1-Score, which reflects a well-balanced performance in precision and recall. To optimize the model, the TensorRT framework was used to accelerate inference for FP32 and to perform post-training quantization for FP16 and INT8 formats. TensorRT's mixed-precision capabilities enabled these optimizations, which reduced the model size, increased inference speed, and lowered power consumption by approximately a factor of two, all while maintaining the original accuracy. Power consumption analysis confirmed that the optimized models used significantly less energy during inference, reinforcing their suitability for deployment in resource-constrained environments. The system was deployed with a Wi-Fi Access Point (AP) hotspot and a web-based interface, enabling users to upload and analyze images directly through connected devices such as mobile phones. This setup ensures simple access and seamless connectivity, making the tool practical for real-world applications. These advancements position the diagnostic tool as an efficient, scalable, and energy-conscious solution to address diagnosis challenges in underserved regions, paving the way for broader adoption in low-resource healthcare settings. 
\end{abstract}


\section{Introduction}
The COVID-19 pandemic exposed severe global inequalities in healthcare accessibility, particularly in developing nations with constrained resources and inadequate infrastructure. At the same time, it highlighted the critical importance of accurate and timely diagnostic tools in controlling infectious disease outbreaks \cite{levin_assessing_2022}. With over seven million deaths and countless cases worldwide \cite{noauthor_covid-19_nodate}, this crisis underscored the urgent need for robust public health systems and the development of innovative diagnostic technologies \cite{noauthor_coronavirus_nodate}. Similarly, the recent resurgence of monkeypox (mpox), including the 2022 outbreak \cite{hraib_outbreak_2022} and the new strain in 2024, which has already resulted in more than 15,600 cases and 537 deaths \cite{noauthor_who_nodate}, have further exposed vulnerabilities in our ability to diagnose and contain emerging infectious diseases quickly. 

The symptoms of mpox, such as fever, headaches, muscle pain, and blistering rashes \cite{noauthor_mpox_nodate}, are often challenging to diagnose accurately, particularly in resource-limited settings where advanced diagnostic tools are not readily available. Traditional diagnosis methods, like the Polymerase Chain Reaction (PCR) test, have limitations in identifying mpox accurately due to the short duration of detectable viral presence in the bloodstream and the disease's varying stages \cite{paniz-mondolfi_evaluation_2023}. Moreover, PCR tests often require several days for processing, and access to such testing is particularly constrained in rural or underserved areas. These factors underscore the need for innovative diagnosis methods that deliver rapid and accurate results, especially in healthcare environments with limited infrastructure.

Artificial Intelligence (AI) has emerged as a transformative tool in healthcare diagnosis, offering the ability to process and analyze complex medical data with both speed and precision \cite{habehh_machine_2021, lindroth_applied_2024}. In the context of mpox, AI-driven models can be particularly promising in resource-limited environments, where access to expert healthcare professionals and sophisticated diagnostic technologies is scarce. By leveraging model compression techniques, it is possible to optimize AI architectures for deployment on edge devices without sacrificing precision and minimizing power consumption. This approach enables the deployment of accurate, fast, and accessible diagnostic tools, meeting the urgent need for reliable mpox testing in vulnerable and underserved populations.

The motivation behind this work is not only to address the immediate diagnostic needs for mpox but also to contribute to the broader advancement of AI-driven diagnostic tools in public health, particularly in underserved regions. By optimizing AI for embedded systems, this research lays the groundwork for further exploration into scalable solutions that can be adapted for other edge devices and healthcare applications. In addition, the methods developed in this research can be expanded to other fields such as agriculture and environmental monitoring, making AI technology more sustainable and accessible across various domains.

In this research, a baseline for binary and multiclass classification of monkeypox skin lesions was created using state-of-the-art deep-learning architectures, including MobileNetV2 \cite{sandler2019mobilenetv2invertedresidualslinear}, EfficientNetB3 \cite{tan2020efficientnetrethinkingmodelscaling}, DenseNet121 \cite{8099726}, InceptionV3 \cite{7780677}, ResNet50 \cite{7780459}, and VGG16 \cite{simonyan2015deepconvolutionalnetworkslargescale}. Then, model optimization was performed utilizing the TensorRT framework, allowing their deployment on a web application hosted on the NVIDIA Jetson Orin Nano. Performance metrics, including model size, average inference time, throughput, and power consumption, were recorded after optimizing FP32 for NVIDIA systems and applying post-training quantization for FP16 and INT8.

Previous studies have used advanced deep learning architectures such as VGGNet \cite{9960194}, ResNet50 \cite{almufareh_transfer_2023}, \cite{Nafisa2022}, EfficientNet \cite{jaradat_automated_2023}, and MobileNetV2 \cite{jaradat_automated_2023}, \cite{Nafisa2023} to make reliable models for diagnosing mpox. These models have demonstrated their effectiveness in medical image classification, offering robust performance for disease detection tasks. However, to the best of the authors' knowledge, this study represents a novel contribution as it not only focuses on the compression and optimization of the model but also actively deploys it on the Jetson Orin Nano.

\section{Methods}

This research investigates the feasibility of an AI-driven diagnostic tool, specifically optimized for embedded systems, to deliver accurate and rapid monkeypox (mpox) diagnosis in resource-limited environments. By leveraging cutting-edge model optimization techniques and maintaining low power consumption, the proposed solution aims to bridge the diagnosis gap in low-resource settings. Additionally, by allowing seamless connection to edge devices through a Wi-Fi Access Point (AP) hotspot, the diagnostic tool ensures accessibility and ease of use, enabling healthcare providers in remote or resource-limited areas to perform rapid and reliable diagnosis without the need for complex infrastructure.

\subsection{Model Training}

This study utilized six deep-learning models: MobileNetV2, EfficientNetB3, DenseNet121, InceptionV3, ResNet50, and VGG16. These models, pre-trained on the ImageNet dataset, were used to implement transfer learning (TL), a technique that applies knowledge from one task or dataset to improve model performance on a related task\cite{noauthor_what_2024}. Specifically, this method involves freezing the majority of the model's layers, preserving the pre-trained weights, and re-training the final layers on the new dataset to adapt the model to the target classification problem. This approach significantly reduces the computational load and training time.

This study utilized two skin lesion datasets: the Monkeypox Skin Lesion Dataset (MSLD) \cite{MSLDV1} and the Mpox Skin Lesion Dataset Version 2.0 (MSLD v2.0) \cite{MSLDV2}. The MSLD dataset is used for binary classification, with two classes: "Monkeypox" and "Others." In contrast, MSLD v2.0 contains six distinct classes: Mpox, Chickenpox, Measles, Cowpox, Hand-foot-mouth disease (HFMD), and Healthy. To address the dataset’s limited size and diversity, data augmentation techniques such as rotation, contrast adjustment, and scaling were applied, incorporating the transformed images into the dataset. These transformations improve model robustness by creating a more representative dataset for training \cite{noauthor_what_2024-1}. The pre- and post-augmentation dataset distribution is shown in Table \ref{tab:datasets}.

\begin{table}[htbp]
\centering
\caption{Original and Augmented Skin Lesion Datasets}
\label{tab:datasets}
\begin{tabular}{|l|l|r|r|}
\hline
\textbf{Dataset} & \textbf{Class} & \textbf{Original} & \textbf{Augmented} \\
\hline
MSLD & Monkeypox & 102 & 1,428 \\
& Others & 126 & 1,764 \\
\cline{2-4}
& \textit{Total} & \textit{228} & \textit{3,192} \\
\hline
MSLD v2.0 & Mpox & 284 & 3,976 \\
& Chickenpox & 75 & 1,050 \\
& Measles & 55 & 770 \\
& Cowpox & 66 & 924 \\
& HFMD & 161 & 2,254 \\
& Healthy & 114 & 1,596 \\
\cline{2-4}
& \textit{Total} & \textit{755} & \textit{10,570} \\
\hline
\end{tabular}
\end{table}

To address the potential for bias caused by the class imbalance in the augmented dataset, 5-fold cross-validation was implemented \cite{noauthor_cross-validation_nodate}. This technique ensured that the data was split into 70\% for training, 20\% for validation, and 10\% for testing, rotating through different subsets to guarantee a comprehensive evaluation of the model’s performance. The details of the hyperparameters utilized in training are highlighted in Table \ref{tab:Hyperparameters}. All training was conducted on an NVIDIA GeForce RTX 4090 GPU.

\begin{table}[htbp]
\caption{Hyperparameters Description}
\begin{center}
\begin{tabular}{|c|c|}
\hline
\textbf{Hyperparameters} & \textbf{Value} \\
\hline
Batch Size & 32 \\
Learning Rate & 0.001 \\
folds & 5 \\
Epochs & 50 \\
Optimizer & Adam \\
Loss & Binary Cross Entropy \\
\hline
\end{tabular}
\label{tab:Hyperparameters}
\end{center}
\end{table}

\subsection{Performance Metrics}
During the training phase, the performance metrics of Accuracy, Precision, Recall, and F1-Score were carefully monitored to ensure the development of a reliable diagnostic tool suitable for real-world deployment. These metrics were chosen to provide a comprehensive evaluation of the model’s performance, with each offering unique insights into its predictive capabilities. Accuracy measures the overall correctness of the model by considering both true positives and true negatives, while Precision focuses on the proportion of correctly identified positive cases among all predicted positives. Recall evaluates the model’s ability to identify all actual positive cases, ensuring minimal false negatives, and F1-Score balances Precision and Recall to provide a single metric reflecting the model's overall performance. The mathematical formulas of these metrics are provided in Eq. 1-4.

\begin{equation}
\text{Accuracy} = \frac{\text{True Positives} + \text{True Negatives}}{\text{Total}}
\end{equation}
\begin{equation}
\text{Precision} = \frac{\text{True Positives}}{\text{True Positives} + \text{False Positives}}
\end{equation}
\begin{equation}
\text{Recall} = \frac{\text{True Positives}}{\text{True Positives} + \text{False Negatives}}
\end{equation}
\begin{equation}
\text{F1-Score} = 2 \cdot \frac{\text{Precision} \cdot \text{Recall}}{\text{Precision} + \text{Recall}}
\end{equation}

A confusion matrix is also employed to evaluate the model's performance during the testing phase. This provides a detailed visualization of the model's predictions, illustrating the number of correct classifications as well as any misclassifications across the categories. By presenting this data in a clear and organized manner, the confusion matrix helps identify potential biases in the model, such as a tendency to favor one class over another. An example of this is given in Table \ref{tab:confusion_matrix}.

\begin{table}[htbp]
\caption{Confusion Matrix}
\begin{center}
\begin{tabular}{c|cc|}
\cline{2-3}
& {\textbf{Predicted Positive}} & {\textbf{Predicted Negative}} \\
\hline
\multicolumn{1}{|c|}{\textbf{Actual Positive}} & True Positive (TP) & False Negative (FN) \\
\hline
\multicolumn{1}{|c|}{\textbf{Actual Negative}} & False Positive (FP) & True Negative (TN) \\
\hline
\end{tabular}
\label{tab:confusion_matrix}
\end{center}
\end{table}

\subsection{Model Compression \& Optimization}

The NVIDIA's TensorRT framework \cite{noauthor_nvidiatensorrt_2024} was utilized in order to compress and optimize the model for deployment on the NVIDIA Jetson Orin Nano, a compact yet powerful embedded system \cite{noauthor_jetson_nodate}. Specifically, TensorRT's post-training quantization capabilities were utilized to optimize our model with mixed precision. By utilizing mixed precision, TensorRT allows different tensors to be quantized to varying levels of numerical accuracy, preventing the metrics from degrading. This produces models optimized for NVIDIA's systems while decreasing model size and improving inference speed and throughput. Currently, TensorRT supports optimizing for TRT-compatible FP32 and post-training quantization for FP16, and INT8 which were used to benchmark the optimization's effectiveness. Model compression and optimization were performed on the Jetson Orin Nano which contains 1024-core NVIDIA Ampere architecture GPU with 32 Tensor Cores.

\subsection{Deployment}

Lastly, the diagnostic tool was deployed via a Wi-Fi AP Hotspot \cite{noauthor_creating_2019} and a web interface, facilitating seamless connectivity for edge devices. In AP mode, the Wi-Fi setup creates a dedicated wireless communication network, allowing the Jetson Orin Nano to function as a gateway for nearby devices, such as smartphones, to connect without requiring an external internet connection. This configuration enables the Jetson Orin Nano to handle wireless security protocols and assign IP addresses to all connected devices, ensuring a secure and stable connection.

To establish the AP, the device must first have a wireless network card. The network is configured with a unique SSID (Service Set Identifier), commonly referred to as the network name, and a secure password to restrict unauthorized access. Once the network is established, the diagnostic tool hosts a web server within this private network. The server is deployed using a public host and port, allowing connected devices to access its interface seamlessly. This setup supports simultaneous connections from multiple devices, enabling users to leverage the diagnosis service efficiently. By integrating this deployment method, the tool ensures broad accessibility and usability, making it an effective solution for resource-limited environments where internet access may be unreliable or unavailable. Power consumption measurements were collected using a P3 International Kill-A-Watt \cite{noauthor_kill--watt_nodate}.


\subsection{Web Interface}

\begin{figure}[htbp]
\centerline{\fbox{\includegraphics[width=0.45\textwidth,height=0.3\textheight,keepaspectratio]{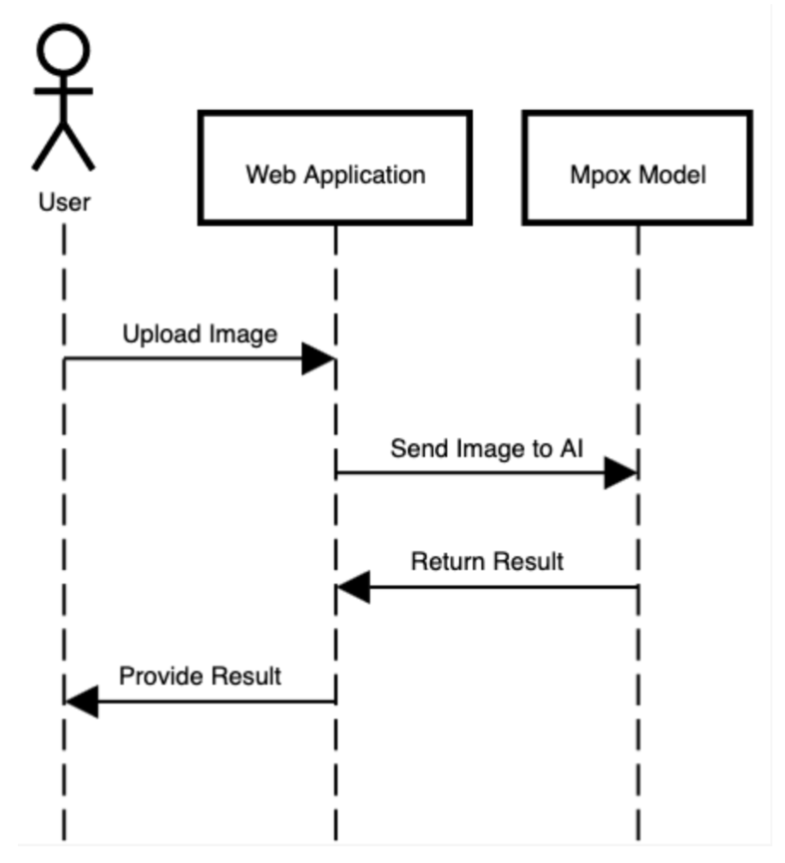}}}
\caption{Web Interface}
\label{fig:interface}
\end{figure}

When connected to the Wi-Fi hotspot via a mobile device, users can access the diagnostic tool's web server by entering the designated port address into their preferred web browser. Upon loading the web application, users are presented with an interface that allows them to either upload an existing image from their device's library or capture a new one directly using their camera. A preview of the selected image is then displayed, and the "Upload" button becomes active. Once the user presses "Upload," the image is submitted to the underlying AI model for processing. The model performs inference and generates a prediction, which is then displayed on a new page. This results page includes the original image preview alongside the model’s classification result. From there, users have the option to return to the home page and repeat the process as needed. The web application provides a straightforward and user-friendly experience, enabling efficient and accurate diagnosis predictions. The web application interface is illustrated in the interaction diagram in Fig.~\ref{fig:interface}.

\section{Results and Discussion}
With the proposed methodology, key performance metrics were collected for both pre-compression and post-compression stages to evaluate the impact of model optimization. These metrics provide a comparative analysis of the model's size, inference speed, and throughput before and after compression. Additionally, the power consumption of the NVIDIA Jetson device was measured in two states: idle and during inference. This analysis was conducted to evaluate the impact of model compression on energy usage and to ascertain the viability of implementing the optimized model in environments with limited resources. 

\subsection{Pre-Optimization}
The performance metrics across all folds were obtained by performing TL and running tests for the MSLD and MSLD v2.0 datasets. These results are shown in Table \ref{tab:MSLD_transfer_learning_results} and Table \ref{tab:MSLD_v2_transfer_learning_results} and collectively underscore the model's robustness and suitability for practical deployment in scenarios demanding high accuracy and balanced performance.

\begin{table}[htbp]
\caption{MSLD Post-Training Results}
\begin{center}
\resizebox{\columnwidth}{!}{
\begin{tabular}{|l|c|c|c|c|}
\hline
Model & Accuracy (\%) & Precision (\%) & Recall (\%) & F1-Score (\%) \\
\hline
MobileNetV2 & $90.75 \pm 1.18$ & $89.56 \pm 2.27$ & $94.35 \pm 1.13$ & $91.87 \pm 0.93$ \\
EfficientNetB3 & $75.66 \pm 15.93$ & $75.12 \pm 15.30$ & $88.02 \pm 7.09$ & $80.73 \pm 11.77$ \\
DenseNet121 & $81.19 \pm 16.14$ & $81.38 \pm 15.36$ & $89.34 \pm 6.04$ & $84.80 \pm 11.17$ \\
InceptionV3 & $83.56 \pm 13.67$ & $84.58 \pm 14.37$ & $89.18 \pm 5.25$ & $86.43 \pm 10.03$ \\
ResNet50 & $80.62 \pm 13.58$ & $81.17 \pm 14.58$ & $88.25 \pm 5.09$ & $84.15 \pm 10.07$ \\
VGG-16 & $79.55 \pm 19.59$ & $80.01 \pm 13.53$ & $87.16 \pm 5.26$ & $83.09 \pm 9.47$ \\
\hline
\end{tabular}
}
\label{tab:MSLD_transfer_learning_results}
\end{center}
\end{table}

\begin{table}[htbp]
\caption{MSLD v2.0 Post-Training Results}
\begin{center}
\resizebox{\columnwidth}{!}{
\begin{tabular}{|l|c|c|c|c|}
\hline
Model & Accuracy (\%) & Precision (\%) & Recall (\%) & F1-Score (\%) \\
\hline
MobileNetV2 & $87.15 \pm 1.88$ & $85.26 \pm 1.95$ & $88.93 \pm 0.55$ & $86.68 \pm 1.47$ \\
EfficientNetB3 & $87.07 \pm 1.79$ & $85.13 \pm 1.93$ & $88.84 \pm 0.54$ & $86.56 \pm 1.44$ \\
DenseNet121 & $ 87.27 \pm 1.67$ & $85.28 \pm 1.79$ & $88.90 \pm 0.65$ & $86.70 \pm 1.39$ \\
InceptionV3 & $87.22 \pm 1.63$ & $85.23 \pm 1.76$ & $88.88 \pm 0.58$ & $86.66 \pm 1.33$ \\
ResNet50 & $87.22 \pm 1.62$ & $85.23 \pm 1.76$ & $88.87 \pm 0.58$ & $86.65 \pm 1.33$ \\
VGG-16 & $87.25 \pm 1.61$ & $85.25 \pm 1.76$ & $88.88 \pm 0.61$ & $86.67 \pm 1.33$ \\
\hline
\end{tabular}
}
\label{tab:MSLD_v2_transfer_learning_results}
\end{center}
\end{table}

\begin{figure}[htbp]
\centerline{\includegraphics[width=0.5\textwidth,keepaspectratio]{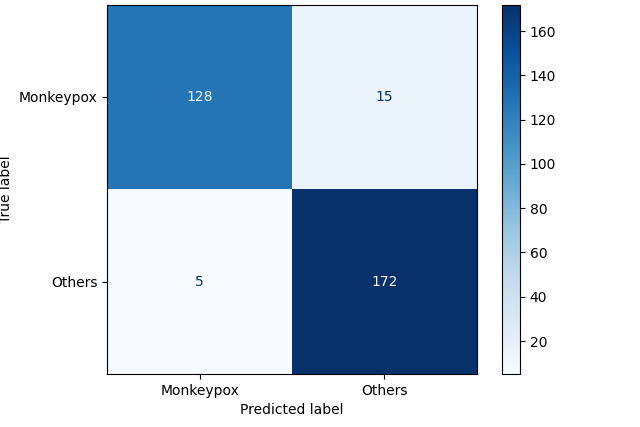}}
\caption{Confusion Matrix for Held-Out Dataset}
\label{fig:confusion_matrix}
\end{figure}

The model demonstrated acceptable performance, achieving an F1-Score of 93.07\%. This metric highlights the model's ability to balance precision and recall effectively, with precision measured at 91.30\% and recall at an even higher 94.92\%. Such results indicate that the model not only correctly identifies true positive cases consistently but also minimizes false negatives, ensuring comprehensive coverage of relevant instances. Furthermore, the overall accuracy of 92.19\% reinforces the model's reliability and consistency in delivering precise predictions across the dataset. The confusion matrix for these results is shown in Fig.~\ref{fig:confusion_matrix}.

The confusion matrix reveals that the model misclassified only 20 images out of the 320 in the test dataset, highlighting its overall robustness and effectiveness in making accurate predictions. However, a closer examination of the misclassifications uncovers a notable limitation: 15 of the 20 misclassified images were incorrectly labeled as "Others" instead of "Monkeypox." This high proportion of false negatives is particularly concerning in a diagnosis context, as failing to identify positive cases of mpox could lead to delays in treatment and increased risks to patient health. False negatives are critical to address because they undermine the reliability of the diagnostic tool in real-world applications, where timely and accurate identification of infectious diseases is essential for effective containment and care. Minimizing these errors is vital to improving the model’s performance and ensuring its safe deployment in clinical and resource-limited environments.

\subsection{Post-Optimization}
In order to effectively evaluate the optimization of the model, three key benchmarks were established, focusing on model size, average inference speed, and throughput. These benchmarks provide a comprehensive framework for assessing the model's efficiency across various dimensions. These tests were conducted on 228 non-augmented images with a batch size of 32.
S
\begin{figure}[htbp]
\centerline{\includegraphics[width=0.5\textwidth,keepaspectratio]{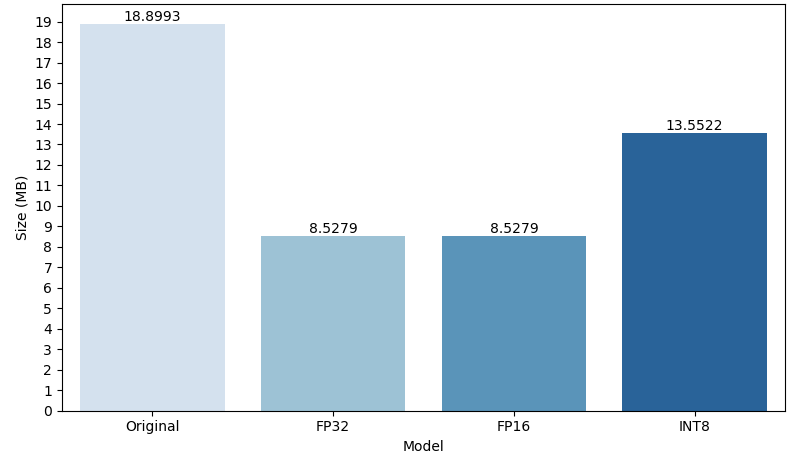}}
\caption{Post-Compression Model Size}
\label{fig:model_size}
\end{figure}

The benchmarks clearly show that TensorRT effectively optimizes the original model. As shown in Fig.~\ref{fig:model_size}, model size is significantly reduced, with decreases by factors of 0.45, 0.45, and 0.72 for FP32, FP16, and INT8 formats, respectively.

\begin{figure}[htbp]
\centerline{\includegraphics[width=0.5\textwidth,keepaspectratio]{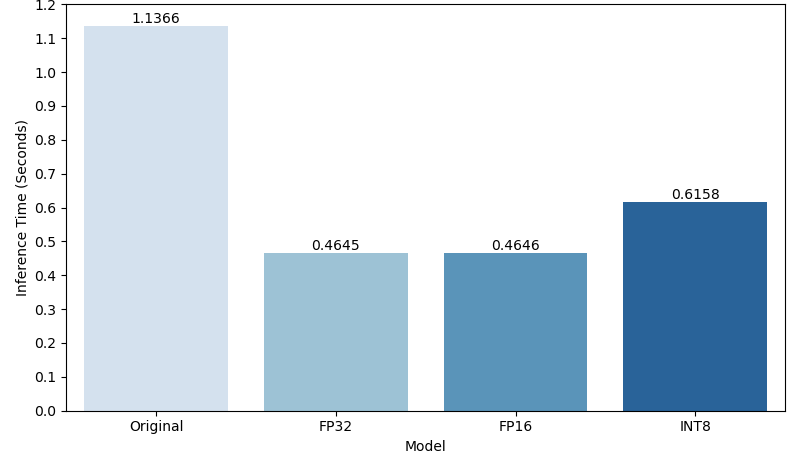}}
\caption{Post-Compression Average Inference Time}
\label{fig:model_average_inference}
\end{figure}

\begin{figure}[htbp]
\centerline{\includegraphics[width=0.5\textwidth,keepaspectratio]{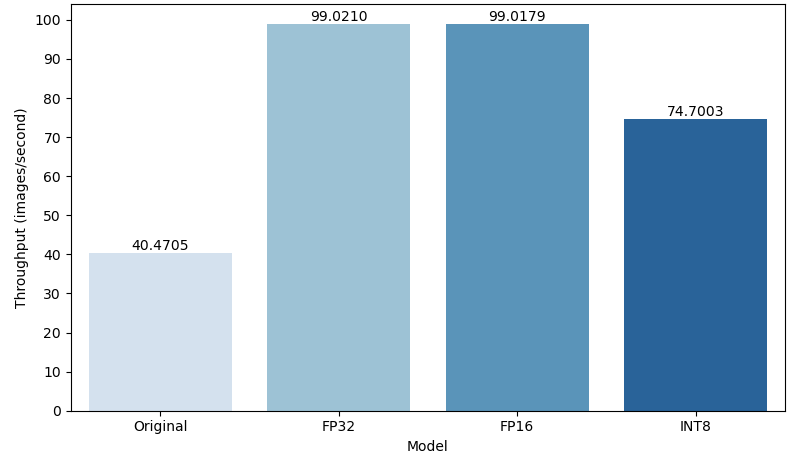}}
\caption{Post-Compression Throughput}
\label{fig:model_throughput}
\end{figure}

Fig.~\ref{fig:model_average_inference} shows the average inference speed improves dramatically compared to the original model, which leads to a corresponding increase in throughput, shown in Fig.~\ref{fig:model_throughput}. An interesting behavior is that FP32 and FP16 demonstrate the same values for average inference time and throughput. The hypothesis for this is that Jetson's hardware limitations do not permit efficient optimization for FP16 precision. Similarly, INT8 demonstrates slightly worse performance than the other precision, most likely due to the 8GB of RAM limiting the calibration process during optimization, which led to worse performance than initially expected. In other words, TensorRT only quantizes certain tensors to INT8, which limits the potential improvements for this format.

\begin{table}[htbp]
\centering
\caption{Comparison of Model Architectures Post-Optimization}
\begin{center}
\begin{tabular}{|l|c|c|}
\hline
\textbf{Operation Type} & \textbf{FP32} & \textbf{FP16/INT8} \\
\hline
AddV2 & 63 & 63 \\
\hline
Cast\textsuperscript{*} & -- & 4 \\
\hline
Const & 125 & 126 \\
\hline
Conv2D & 35 & 35 \\
\hline
DepthwiseConv2dNative & 17 & 17 \\
\hline
MatMul & 1 & 1 \\
\hline
Mean & 1 & 1 \\
\hline
Mul & 17 & 18 \\
\hline
Pad & 4 & 4 \\
\hline
Relu6 & 35 & 35 \\
\hline
\end{tabular}
\label{tab:model_architectures}
\end{center}
\textit{\textsuperscript{*}Cast operations are only present in FP16/INT8 models for precision conversion}
\end{table}

Table \ref{tab:model_architectures} presents the optimized model architectures obtained after applying optimization techniques. The original MobileNetV2 architecture has been modified to enhance performance specifically for the NVIDIA Jetson Orin Nano. TensorRT optimizes models using mixed-precision techniques, which involve retaining the original precision (e.g., FP32) for input and output layers, while optimizing the precision of hidden layers. These optimizations typically include fusing multiple layers into single operations, adding constant tensors, and optimizing certain operations such as Depthwise Convolution. These changes reduce the overall number of operations while maintaining the model’s accuracy. For FP16 and INT8 precision modes, the Cast operation is introduced to handle conversions between the input in FP32 and the target precisions. It is also important to note that some operations, such as NoOp, Identity, and Placeholder, are not supported by TensorRT and remain unchanged during the optimization.

\subsection{Power Consumption}
Table \ref{tab:power_consumption} displays the power consumption data for the Jetson Orin Nano. This provides valuable insight into the energy efficiency of the device, highlighting how its power usage varies with the different optimized models and how it compares to the original. 

\begin{table}[htbp]
\caption{Jetson Orin Nano Power Consumption}
\begin{center}
\resizebox{\columnwidth}{!}{
\begin{tabular}{lcc}
\toprule
Model Type & Inference Consumption (W) \\
\midrule
Original Model & $\sim$6.0 \\
FP32 & $\sim$5.5 \\
FP16 & $\sim$5.5 \\
INT8 & $\sim$5.4 \\
\bottomrule
\end{tabular}
}
\label{tab:power_consumption}
\end{center}
\textit{Note: The Idle Power Consumption was $\sim$5.2 W}
\end{table}

An analysis of power consumption reveals that the optimized models consume significantly less power compared to the original model. This improvement is achieved through optimizing the model, which simplifies the architecture using FP32, FP16, and INT8 precision formats. Specifically, power consumption is reduced by ratios of 0.92, 0.92, and 0.90 for FP32, FP16, and INT8, respectively. These results strongly support the initial hypothesis, demonstrating that TensorRT's post-training quantization and optimization is an effective strategy for minimizing power usage on the Jetson Orin Nano. 

\subsection{Web Application Inference}

\begin{figure}[htbp]
\centerline{\fbox{\includegraphics[width=0.5\textwidth,height=0.3\textheight,keepaspectratio]{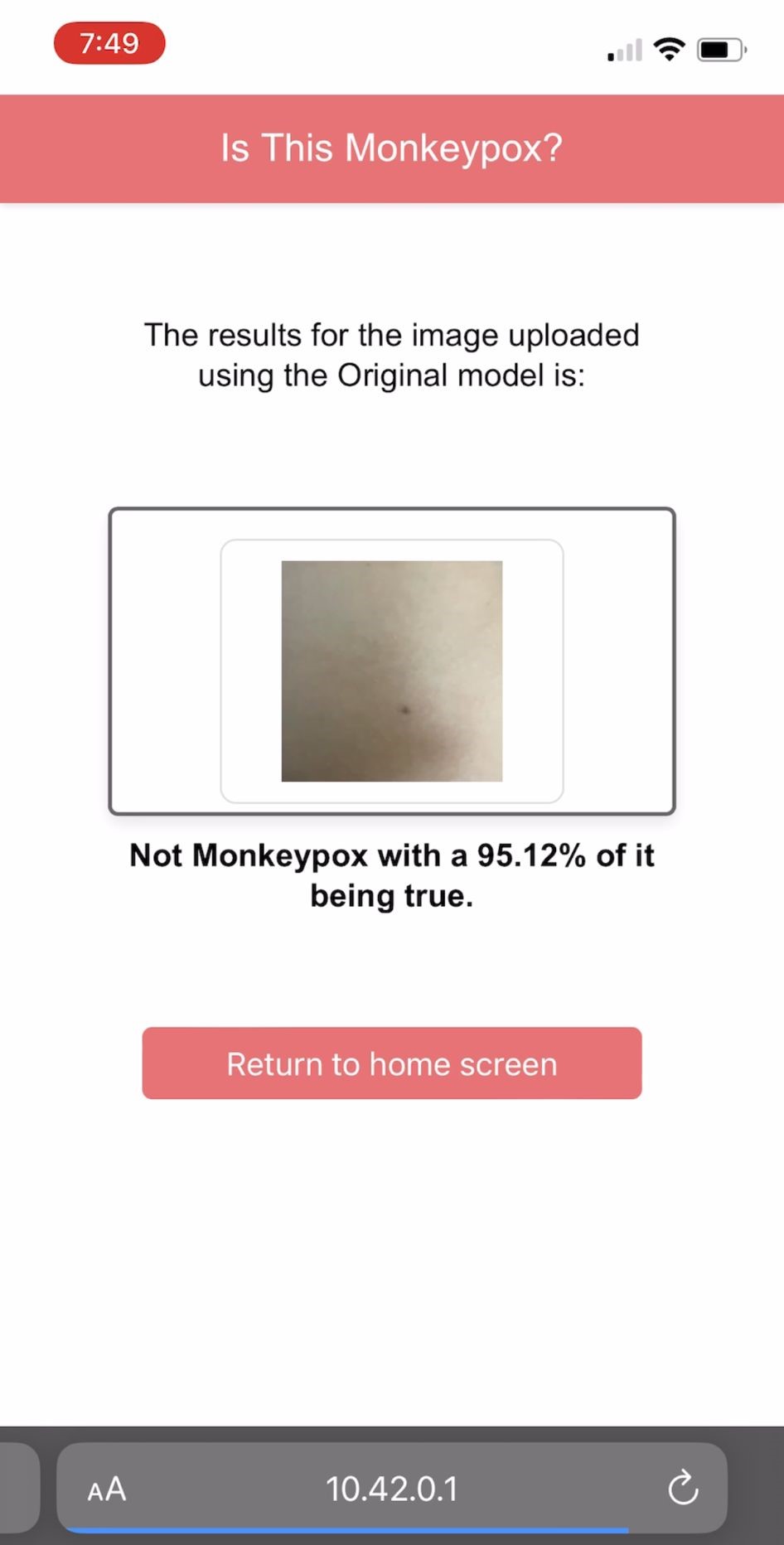}}}
\caption{Web Application Inference Example}
\label{fig:web_app_inference}
\end{figure}

The interface guides the user through a straightforward process: first, it prompts the user to select a model type and upload an image. Once the image is uploaded, the system performs inference, leveraging the MobileNetV2 model to analyze the input. The application then displays the diagnosis results, presenting the original image alongside a prediction indicating whether the lesion is likely to be mpox. Additionally, it provides a confidence score expressed as a percentage, offering users clarity and assurance about the system's decision as shown in Fig.~\ref{fig:web_app_inference} with the original model.

\section{Conclusion}
This study presents a Monkeypox diagnostic tool deployed on the NVIDIA Jetson Orin Nano, utilizing the pre-trained MobileNetV2 architecture for binary classification. The model was trained on the Monkeypox Skin Lesion Dataset, achieving an F1-Score of 93.07\%, which demonstrates an effective balance between precision and recall. Post-training quantization using the TensorRT framework yielded substantial improvements in model size, inference time, and throughput while preserving the original performance metrics. TensorRT's mixed-precision capabilities enabled these enhancements, optimizing the model for deployment without compromising the performance metrics. Additionally, power consumption measurements showed that the optimized models significantly reduced energy usage compared to the original, showcasing their suitability for resource-constrained environments. Lastly, by deploying the model on the Jetson Orin Nano, this research bridges the gap between high-performance medical image analysis and real-world usability, demonstrating a significant advancement in edge-based diagnosis systems.

Looking ahead, future work will explore expanding the base model for transfer learning to include architectures such as EfficientNet and ResNet, among others. Furthermore, the use of more extensive datasets will enable testing the model’s ability to generalize effectively to unseen data, ensuring its robustness and applicability across diverse real-world scenarios. Lastly, deployment on other embedded systems will be explored, such as the Raspberry Pi and Arduino Nano while leveraging other optimization frameworks such as LiteRT \cite{noauthor_litert_nodate} and Apache TVM \cite{noauthor_apache_nodate}.

\section*{Acknowledgment}

This work would not have been possible without the support of the Center for Research \& Development at the University of Puerto Rico at Mayagüez. A sincere thank you to Luis E. Fernandez Ramirez for their assistance throughout the research. Lastly, thank you to the NSF Grant OIA-1849243 for partially funding this research.






\nocite{*}  
\bibliographystyle{IEEEtran}
\bibliography{references}

\end{document}